\documentclass[letterpaper, 10 pt, journal, twoside]{IEEEtran}
\usepackage{amsmath,amsfonts}
\usepackage{stix}
\usepackage{algorithmic}
\usepackage{algorithm}
\usepackage{array}
\usepackage[caption=false,font=normalsize,labelfont=sf,textfont=sf]{subfig}
\usepackage{textcomp}
\usepackage{stfloats}
\usepackage{url}
\usepackage{verbatim}
\usepackage{graphicx}
\usepackage{cite}
\usepackage[table]{xcolor}
\usepackage{booktabs}
\usepackage{enumitem}
\usepackage{multirow}
\usepackage{hyperref}
\usepackage{graphicx} 
\renewcommand{\baselinestretch}{0.982}
\usepackage{caption}
\usepackage[font={scriptsize}]{caption}
\newcommand{\smallerpm}[1]{\text{\scriptsize $\pm$}{\text{\scriptsize #1}}}
\usepackage{float}

\newcommand{\bestres}[1]{\cellcolor{green!25}\textbf{#1}}
\newcommand{\secondbest}[1]{\cellcolor{green!10} #1}

\usepackage{soul} 
\usepackage[normalem]{ulem} 

\begin{document}
%
\title{FrontierNet: Learning Visual Cues to Explore}
%
%
%

\author{Boyang Sun$^{1}$, Hanzhi Chen$^{2}$, Stefan Leutenegger$^{2,3}$, Cesar Cadena$^{4}$, Marc Pollefeys$^{1,5}$, and Hermann Blum$^{1,6}$%
\thanks{Manuscript received: December 19, 2024; Revised: March 25, 2025; Accepted: April 30, 2025. This paper was recommended for publication Editor Abhinav Valada upon evaluation of the Associate Editor and Reviewers' comments. This work was partially supported by the DSO National Laboratories, DSOCO24035.}
\thanks{$^{1}$Boyang Sun, Marc Pollefeys, and Hermann Blum are with Computer Vision and Geometry Group, ETH Zurich, 8092 Zurich, Switzerland (e-mail: boyang.sun@inf.ethz.ch; marc.pollefeys@inf.ethz.ch).}
        
        
\thanks{$^{2} $Hanzhi Chen and Stefan Leutenegger are with the Mobile Robotics Lab, Technical University of Munich, 80333 München, Germany (e-mail: hanzhi.chen@tum.de; stefan.leutenegger@tum.de)}
\thanks{$^{3} $Stefan Leutenegger is also with Mobile Robotics Lab, ETH Zurich, 8092 Zurich, Switzerland (e-mail: lestefan@ethz.ch)}

\thanks{$^{4} $Cesar Cadena is with Robotic Systems Lab, ETH Zurich, 8092 Zurich, Switzerland (e-mail: cesarc@ethz.ch)}

\thanks{$^{5} $Marc Pollefeys is also with Microsoft Mixed Reality and AI Lab, 8038 Zurich, Switzerland (e-mail: mapoll@microsoft.com)}

\thanks{$^{6} $Hermann Blum is also with 
Robot Perception and Learning Lab, University of Bonn and Lamarr Institute for ML and AI, 53115 Bonn, Germany (e-mail: blumh@uni-bonn.de)}

\thanks{Digital Object Identifier (DOI): see top of this page.}
}
%
%

\markboth{IEEE Robotics and Automation Letters. Preprint Version. Accepted MAY, 2025}
{Sun \MakeLowercase{\textit{et al.}}: FrontierNet: Learning Visual Cues to Explore} 

%



\maketitle

\begin{abstract}
Exploration of unknown environments is crucial for autonomous robots; it allows them to actively reason and decide on what new data to acquire for different tasks, such as mapping, object discovery, and environmental assessment. Existing solutions, such as frontier-based exploration approaches, rely heavily on 3D map operations, which are limited by map quality and, more critically, often overlook valuable context from visual cues. This work aims at leveraging 2D visual cues for efficient autonomous exploration, addressing the limitations of extracting goal poses from a 3D map. We propose a visual-only frontier-based exploration system, with \textbf{FrontierNet} as its core component. FrontierNet is a learning-based model that (i) proposes frontiers, and (ii) predicts their information gain, from posed RGB images enhanced by monocular depth priors. Our approach provides an alternative to existing 3D-dependent goal-extraction approaches, achieving a 15\% improvement in early-stage exploration efficiency, as validated through extensive simulations and real-world experiments. The project is available at \href{https://github.com/cvg/FrontierNet}{https://github.com/cvg/FrontierNet}.
\end{abstract}
\vspace{-1mm}

%
\IEEEpeerreviewmaketitle

\section{Introduction}
\label{sec:intro}

\IEEEPARstart{A}{utonomous} exploration requires a robot to navigate through an unknown environment to accomplish tasks such as building a digital map, locating objects, and, more generally, gathering information. This capability is critical for a wide range of applications, including infrastructure modeling and inspection \cite{ginting2024semantic,liu2014infrastructure}, search and rescue \cite{ziparo2013exploration, chen20243d}, crop monitoring \cite{gao2024aerial, lobefaro2023estimating}, and object search \cite{batra2020objectnav}.

Efficient autonomous exploration, whether aimed at maximizing mapped volume, enriching semantic understanding, or boosting reconstruction quality, ultimately boils down to identifying optimal poses for the robot to reach. Existing methods, often based on the 3D map constructed by the robot, either focus on extracting the map boundary \cite{yamauchi1997frontier} or iteratively sample poses or paths within the map and select the most suitable ones \cite{bircher2016receding}. These approaches differ in perspective: one derives poses from the 3D map by calculating optimal poses directly, the other samples poses and evaluates them against the map to find the optimal ones. Thus, both approaches leverage the 3D map information to guide exploration. At the same time, they are also inherently limited by the quality of the 3D map, which depends on factors like sensor accuracy, reconstruction methods, and map representation. More importantly, they tend to overlook the rich appearance cues streaming from the robot’s RGB cameras, such as texture, color, and semantic context, resulting in redundant and inefficient exploration paths.

\begin{figure}[t] 
    \centering
    \includegraphics[width=0.485\textwidth]{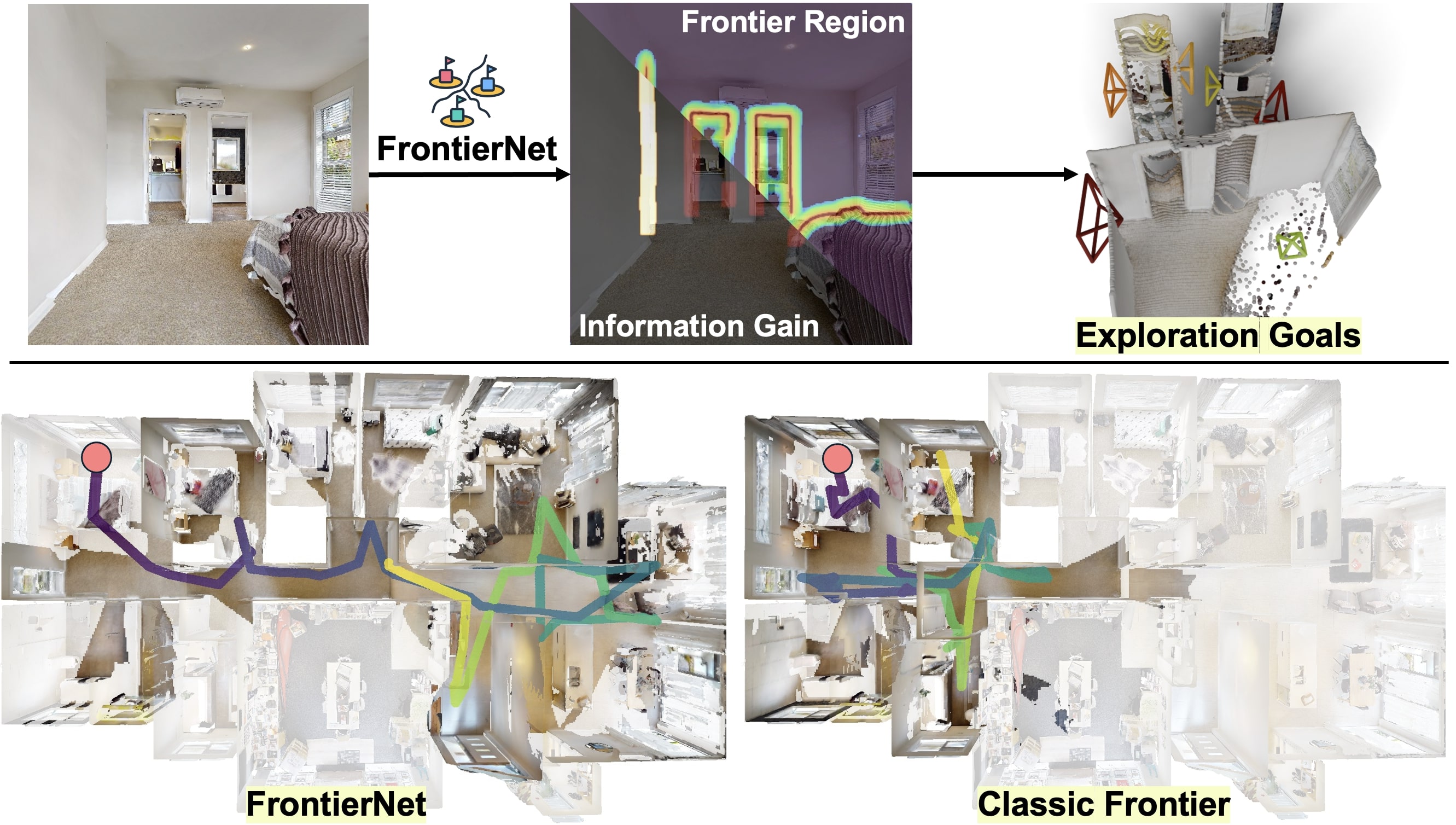}
    \caption{\textbf{Top}: FrontierNet processes a RGB image (left) to propose frontier pixels and their information gain (middle), registering candidate goal viewpoints with varying priorities in 3D (right). \textbf{Bottom}: Using FrontierNet, our exploration system prioritizes visiting unknown regions with greater potential of unmapped volume, achieving higher efficiency.}
    \label{fig:teaser}
    \vspace{-1.5\baselineskip}
\end{figure}

\begin{figure}[t] 
    \centering
    \includegraphics[width=0.48\textwidth]{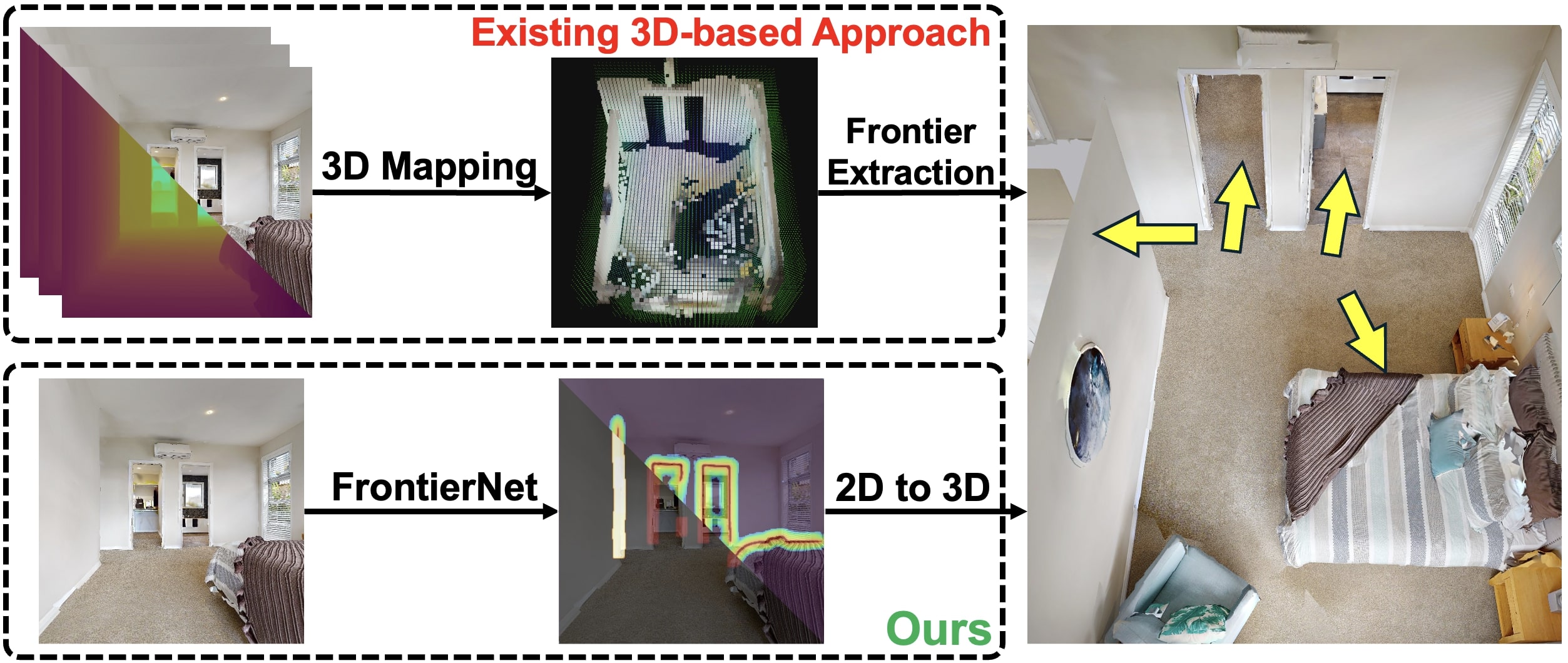}
\vspace{-1mm}  \caption{FrontierNet learns to propose regions for exploration from visual cues in RGB images. Unlike existing methods, it avoids operations on dense 3D maps at the proposal stage, which are sensitive to map quality, and often discard rich appearance information.}
    \label{fig:intro}
    \vspace{-1.5\baselineskip}
\end{figure}

In contrast to \textbf{dense} 3D map operations typically used in exploration, the final solution to exploration often results in \textbf{sparse} outputs, such as a set of goal poses. Sparse representations like these have proven effective and efficient for various robotic tasks, including exploration and navigation \cite{yokoyama2024vlfm,schmid2020efficient,dai2020fast,papatheodorou2023finding,zhou2021fuel,tao2023seer,chaplot2018active,cao2021tare,papatheodorou2024efficient}. We argue that achieving similarly sparse outputs does not inherently require dense 3D operation. For instance, a human can readily identify key spots to move to uncover unknown spaces from a single RGB image. These spots, which represent the explicit boundary of the current viewpoint, are akin to 3D map boundary but can be inferred with visual-only input. 
This inference relies solely on cues from RGB images, while effectively extracting both geometric and appearance information. Additionally, one can estimate how much unknown space each spot might reveal, informed by contextual image details—a level of inference that is challenging and costly in 3D. Fig. \ref{fig:intro} provides an abstract comparison of identifying candidate poses for exploration using visual cues versus dense 3D geometry.

Building on these observations, this work explores how to extract explicit boundary indicators from RGB images for autonomous exploration. We propose a visual-only frontier-based exploration approach, introducing FrontierNet, a learning-based model for hybrid frontier proposal and information gain prediction. This model directly proposes frontiers and predict their information gain from individual RGB frames, linking exploration decisions in 3D space with 2D visual cues. Our system supports posed RGB input and augments it with monocular depth priors. The contributions of this paper are summarized as follows:
\begin{itemize}[leftmargin=*]
    \item An efficient autonomous exploration system that exploits visual cues available in individual camera images.
    \item A learning-based frontier proposal and information gain prediction model integrated in the proposed system. 
    \item Extensive simulation experiments and real-world tests that validate the model and the proposed system.
\end{itemize}

\section{Related Work}

Various approaches have been proposed for autonomous exploration. As introduced in Section \ref{sec:intro}, two major types: \textit{frontier-based} and \textit{sampling-based} methods are commonly used to solve the problem. Most of these methods rely on a 3D representation of the world to operate on. They have different objectives and represent the environment in distinct ways. Early works use conventional 3D representations, such as occupancy grid \cite{yamauchi1997frontier,gao2018improved,selin2019efficient}, signed distance field \cite{bircher2018receding, schmid2020efficient} and 3D point cloud \cite{cao2021tare}, with which frontier-based methods iterate through the map and extract the map boundary, while sampling-based methods evaluate sampled viewpoints using different metrics, such as map entropy and uncertainty \cite{dai2020fast,schmid2020efficient}. More recent work has tried to use learning-based vision algorithms to help design evaluation metrics. In \cite{tao2023seer}, a 3D occupancy prediction model is used to estimate the information gain of each frontier. \cite{schmid2022sc} uses similar scene completion network for viewpoint evaluation. With the emerging new 3D representations, recent works have proposed the use of neural implicit representation \cite{Yan2023iccv, lee2022uncertainty}, or 3D Gaussian \cite{Jiang2023FisherRF,jiang2024ag,tao2024rt}.

The aforementioned works have shown that 3D geometry representation can be helpful for exploration; recent approaches build on this by incorporating appearance information into the 3D representation for improved performance. One line of work introduces object-level semantics into the maps, \cite{asgharivaskasi2021active, papatheodorou2023finding,tao2023seer} introduce semantic information into trajectory and viewpoint evaluation, and \cite{tao20243d} uses semantic-informed loop closure for better localization accuracy during exploration. Another branch of work model exploration as a decision-making problem, they use reinforcement learning to solve the problem that often includes the color image as input \cite{chaplot2018active,chaplot2021seal}. More recent works try to utilize the power of vision foundation models and large language models for interactive, human-like exploration \cite{chen2023not, yokoyama2024vlfm, qu2024ippon}.

The mentioned works have shown that appearance is a valuable resource for exploration. Although appearance information has been utilized, it is either tightly integrated with volumetric maps for metric design or serves as input for independent vision algorithms. However, we observe that appearance cues can be directly leveraged when identifying boundaries without relying on 3D representations. These cues also allow for the evaluation of boundaries, eliminating the need to integrate them into intermediate visual task models.

\section{Method}
\subsection{Problem Statement}
\label{subsec:pre}
The goal of this work is to let a camera-equipped robot autonomously explore an environment. As it moves, the robot continuously captures images and leverages them to expand and refine its knowledge of the environment. To quantify this knowledge, we follow prior works \cite{zhou2021fuel,tao2023seer,schmid2020efficient} and choose mapped volume as the metric. A static environment can be modeled as a bounded volume $\mathbf{V} \subset \mathbb{R}^3$, each point $\mathbf{v} \in \mathbf{V}$ is associated with occupancy probability $P({\mathbf{v}})$. Initially, all the points have $P({\mathbf{v}})=0.5$, indicating occupancy as \textit{unknown}. The occupancy probability of each point gets updated when the robot extends its map covering it. It becomes a \textit{known} point, i.e., $\mathbf{v} \in \mathbf{V}_{\text{known}}$, where $\mathbf{V}_{\text{known}} \subset \mathbf{V}$. We aim to find a sequence of poses $\mathbf{x} = (\mathbf{p}, \mathbf{q}) $, $\mathbf{p} \in \mathbb{R}^3$ and $\mathbf{q} \in \mathbb{SO}(3)$, which the robot follows and collects images to maximize $|\mathbf{V}_{\text{known}}|$.
 
\subsection{System Overview}

\begin{figure}[!t] 
    \centering
    \includegraphics[width=0.45\textwidth]{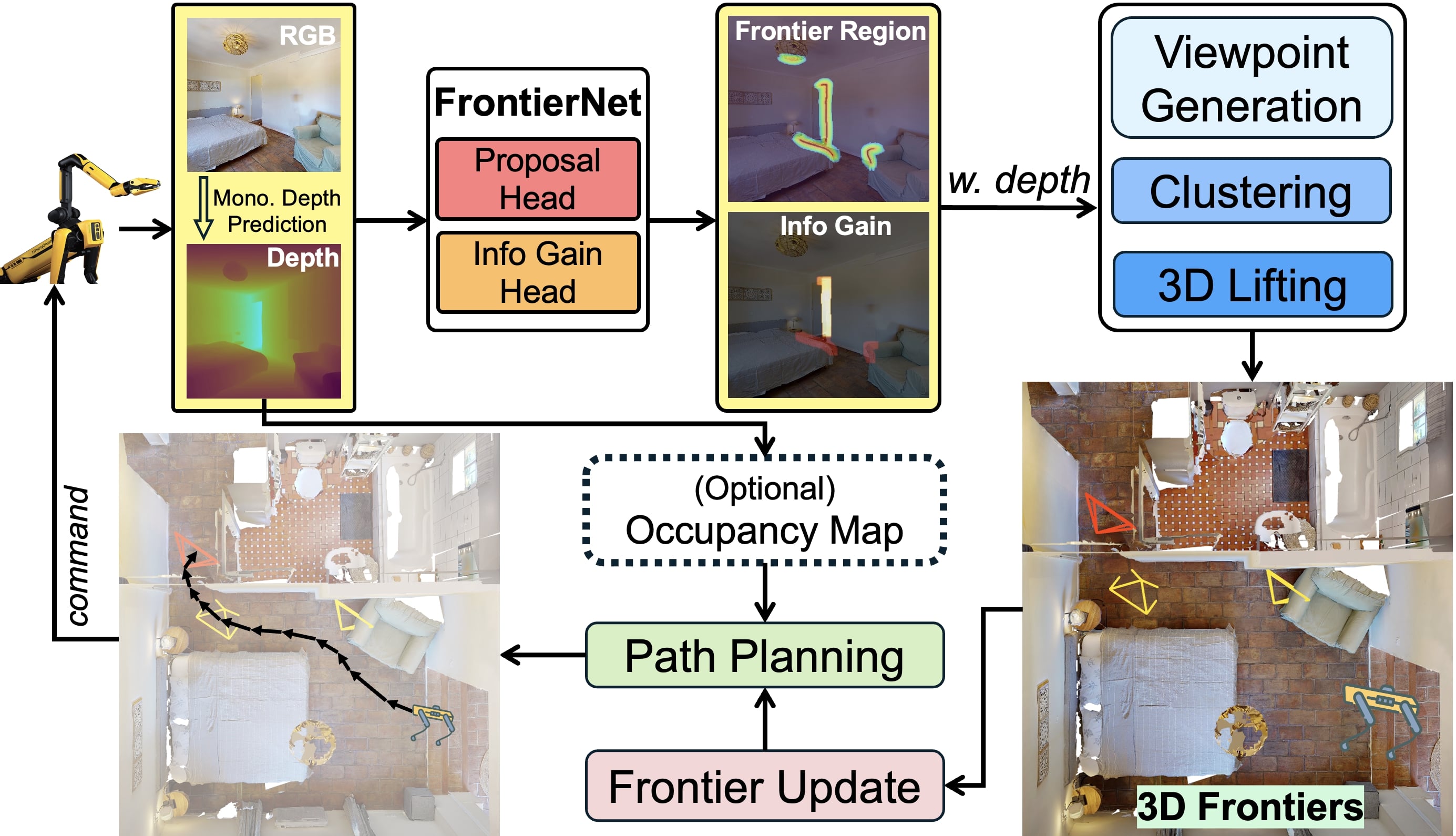}
\vspace{-1mm}
\caption{\textbf{System Overview. }Our system processes posed RGB images with a depth prediction model \cite{hu2024metric3d} to generate estimated depth. FrontierNet uses visual input to predict 2D frontier regions and their info gain, which are transformed into sparse 3D frontiers with different gains (colored frustums). These frontiers are tracked, and the planning module selects the next best goal and plans a path using the occupancy map. }
    \label{fig:system}
    \vspace{-1.0\baselineskip}
\end{figure}

An overview diagram of the proposed system can be seen in Fig. \ref{fig:system}. The core component is our FrontierNet, which performs joint frontier proposal and information gain prediction, followed by 3D-anchoring and planning steps. During exploration, our system maintains a frontier updating mechanism that tracks changes across all frontiers. The path planning module selects the next goal frontier and plans a path. 

\subsection{\texorpdfstring{Learning to Propose Frontiers from Visual Appearance}{Learning to Propose Frontiers from Visual Appearance}}

\label{subsec:frontiernet}

Following Yamauchi’s formulation \cite{yamauchi1997frontier}, we define a frontier as a region of free space that directly borders unexplored space.
Commonly, frontiers are therefore proposed from 3D voxel maps. Instead, we consider \textbf{\textit{frontier pixels}} as the 2D projection of 3D frontier voxels within a camera's observed space and train a model that locates these pixels directly on image plane. Conventional frontier definitions treat every frontier as equally valuable and overlook differences in how much additional space each one can reveal. Recent studies \cite{tao2023seer,schmid2022sc,dai2020fast} address this limitation by introducing quantitative metrics, often called information gain, that rank frontiers according to their expected exploratory benefit. In this work, we define the additional observable volume previously unknown from a frontier as its information gain (\textbf{\textit{info gain}}) and train our model to also predict it from the visual input. This prediction depends only on individual images, assuming no prior exploration. 

To unify the proposal of frontier pixels with the prediction of info gain, we employ a two-head UNet-like structure, \textbf{FrontierNet}, and frame the task as an image-to-image prediction. It utilizes both the color image and its corresponding monocular depth prior as input and jointly predicts the frontier pixels and info gain. 

For the frontier pixels proposal head, inspired by recent advances in line detection \cite{pautrat2023deeplsd, xue2020holistically}, our approach models the frontier pixels using a distance field $\mathbf{D}$. 
Given an input RGB image with its monocular depth prior \( \mathbf{I} \in \mathbb{R}^{H \times W \times 4} \), FrontierNet $f_{\text{FtNet}}(\cdot)$ predicts \( \mathbf{D} \in \mathbb{R}^{H \times W} \), where the value of each pixel \( (i, j) \) in \( \mathbf{D} \) is the distance on the image plane to the closest frontier pixel:
\begin{align}
\label{eq:1}
\tilde{\mathbf{D}} &= f_{\text{FtNet}}(\mathbf{I}), \\
\mathbf{D}[i, j] &= \min_{(x, y) \in \mathcal{F}} \| (i, j) - (x, y) \|_2,
\end{align}
where \(\tilde{\mathbf{D}}\) is the prediction, \( \mathcal{F} \) denotes pixel set corresponding to the frontier pixels in \( \mathbf{I} \), and \( \| \cdot \|_2 \) is the Euclidean distance.
  
For the info gain prediction head, following our definition, the projected 3D voxels with their calculated info gain form a 2D info gain value map $\mathbf{G}  \in \mathbb{R}^{H \times W}$. The calculation of $\mathbf{G}$ will be discussed in \ref{subsec:train}. Regressing the pixel-wise value with high variance can be challenging and sensitive to noisy input\cite{bhat2021adabins}, we reformulate info gain prediction as a multi-class classification problem. We discretize the value spectrum of the info gain into $K$ bins and let the model predict the bin index. Given the input image \( \mathbf{I} \), our model predicts the multi-class info gain map \( \mathbf{Y} \in \mathbb{N}^{H \times W} \) as:
\begin{align}
    \tilde{\mathbf{Y}} &= f_{\text{FtNet}}(\mathbf{I}), \label{eq:2} \\
    \mathbf{Y}[i, j] &= \operatorname{bin}(\mathbf{G}[i, j], K),\label{eq:bin}
\end{align}
where \(\tilde{\mathbf{Y}}\) is the prediction, \( \mathbf{G}[i, j] \) is the info gain at pixel \( (i, j) \), and \( \operatorname{bin}(\cdot, K) \) maps \( \mathbf{G}[i, j] \) into one of \( K \) discrete classes.

\subsection{Data Generation and Model Training}
\label{subsec:train}

\begin{figure}
\centering
  \centering
  \includegraphics[width=0.9\linewidth]{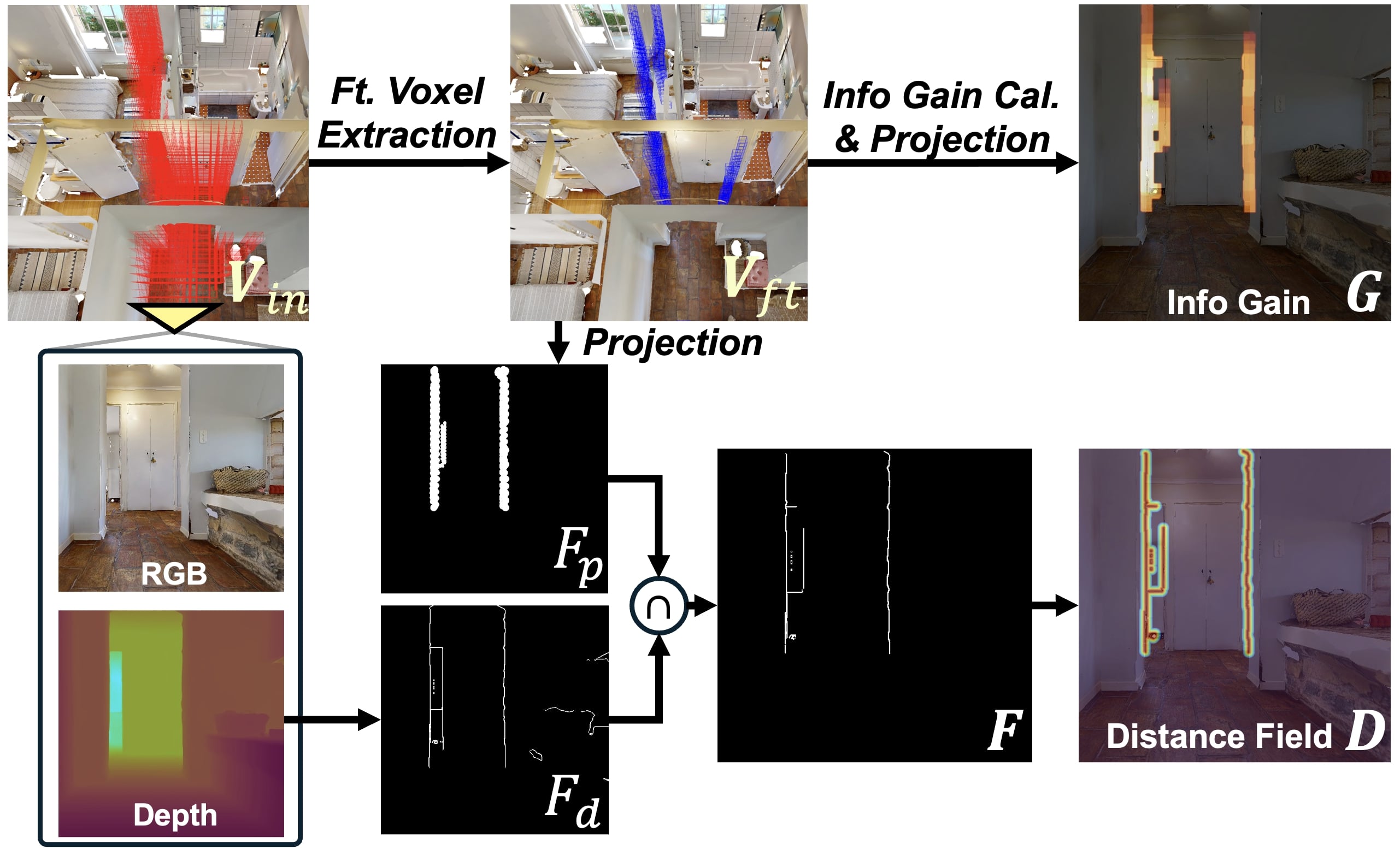}
\vspace{-.5mm}
\caption{\textbf{Ground Truth Generation. }For a sampled camera pose in the voxelized scene, 3D frontier voxels are calculated and projected onto the camera frame using ground truth 3D occupancy grid. Merging the projection with the depth discontinuity mask produces a refined and less noisy frontier pixels mask \( \mathbf{F} \), which is used to calculate the distance field map \( \mathbf{D} \). Additionally, projecting the info gain of each frontier voxel onto the camera frame generates the info gain map \( \mathbf{G} \). }
\label{fig:gt_gen}
\vspace{-7mm}
\end{figure}

Few works have explored learning to propose frontiers or predict information gain directly from images. Some studies leverage intermediate vision modules to estimate information in unknown space, such as map completion approaches \cite{schmid2022sc,tao2023seer}, which take a 3D map as input and hallucinate unknown areas, and then an information gain can be computed. We use 3D information to generate ground truth data and directly supervise our model $f_{\text{FtNet}}(\cdot)$ without intermediate steps. Specifically, we generate ground truth data from Habitat-Matterport 3D (HM3D) \cite{ramakrishnan2021habitat}, a dataset of real-world textured 3D scans.

Fig. \ref{fig:gt_gen} illustrates the ground truth generation pipeline. We voxelize the entire 3D scene and sample camera viewpoints within the voxelized space. The voxel grid is categorized into two classes: voxels inside the camera view (\( \mathbf{V}_{\text{in}} \)) and those outside (\( \mathbf{V}_{\text{out}} \)). Following the logic of the conventional 3D frontier proposal, frontier voxels (\( \mathbf{V}_{\text{ft}} \)) are identified within \( \mathbf{V}_{\text{in}} \) as those adjacent to \( \mathbf{V}_{\text{out}} \). \( \mathbf{V}_{\text{ft}} \) are projected onto the image plane to generate a binary prior \( \mathbf{F}_\text{p} \), representing the initial frontier pixel. Since frontier pixels are typically associated with gaps in appearance and geometry, which often correspond to depth discontinuities, we create a binary depth discontinuity mask \( \mathbf{F}_\text{d} \) by thresholding the depth gradient map. The refined frontier pixels mask \(  \mathbf{F} \) is obtained by intersecting \( \mathbf{F}_\text{p} \) and \( \textbf{F}_\text{d} \), i.e., \( \mathbf{F} = \mathbf{F}_\text{p} \cap \mathbf{F}_\text{d} \). Finally, we generate the ground truth truncated distance field \( \mathbf{D} \) from \( \mathbf{F} \).

To obtain the ground truth info gain, we calculate the additional observable volume for each frontier voxel \( \mathbf{v} \in \mathbf{V}_{\text{ft}} \), and propagate this value to each frontier pixel. Essentially, this uses privileged information to build a dataset from which the model learns correlations between visual appearance and info gain. Ideally, this would involve checking every \( \mathbf{v} \in \mathbf{V}_{\text{ft}} \) and identifying the viewpoint that maximizes observable volume from \( \mathbf{V}_{\text{out}} \); however, this operation is computationally intractable. We approximate this by sub-sampling 10\% of \( \mathbf{V}_{\text{ft}} \). For each sampled voxel, we determine an optimal viewpoint by calculating the 3D direction from \( \mathbf{V}_{\text{in}} \) to \( \mathbf{V}_{\text{out}} \) at its location. We then linearly interpolate the estimated info gain values of the remaining frontier voxels in \( \mathbf{V}_{\text{ft}} \). This approximation is reasonable because (i) at any frontier voxel, the optimal viewing direction to observe unknown space is generally toward regions outside the observed area, and (ii) frontier voxels that are spatially close are also close to the same unknown regions, therefore providing similar info gain. In practice, we generate both \( \mathbf{F}_\text{p} \) and \( \mathbf{G} \) by performing per-pixel ray-casting. For each ray, we compute its distance to all voxels from \( \mathbf{V}_{\text{ft}} \) and retain only those within a specified range \( r \), effectively controlling the extent of the info gain map. The info gain value of the pixel is assigned as the maximum info gain from all voxels close enough to the ray.

We train both heads of FrontierNet simultaneously. One head regresses the distance field \( \mathbf{D} \), while the other classifies the multi-class info gain mask \( \mathbf{Y} \). The input image \( \mathbf{I} \) is processed by a shared encoder-decoder structure based on a ResNet \cite{he2016deep} backbone pretrained on ImageNet \cite{russakovsky2015imagenet}. The shared output is then passed to two separate heads, each consisting of three 2D convolution layers.
To supervise the distance field \( \mathbf{D} \), we apply a normalization process similar to \cite{pautrat2023deeplsd}:
 $ \hat{\mathbf{D}} = -\log \left( \mathbf{D} /r \right)$
For the info gain classification head, we discretize the info gain values into 11 ($K=11$ in Eq. \ref{eq:bin}) classes. 
The total loss is the weighted sum of the two heads:
\begin{align}
    \mathcal{L} =  \alpha \cdot \mathcal{L}_{\text{D}}( \tilde{\mathbf{D}}, \hat{\mathbf{D}})+ \mathcal{L}_{\text{Y}} (\tilde{\mathbf{Y}}, \mathbf{Y}),
\end{align}
where \( \mathcal{L}_{\text{D}} \) is the $\operatorname{L1}$ loss on the normalized distance field, \( \mathcal{L}_{\text{Y}} \) is the combined cross entropy and multi-class Dice loss on the multi-class map, and \( \alpha \) is a hyper-parameter.

\subsection{Anchoring Frontier in 3D} 
We design an anchoring stage that extracts sparse candidate frontiers with viewing directions from the output of FrontierNet and lift them to 3D as targets for the robot to approach. As an initial step, it recovers the frontier pixels and info gain value map \( (\mathbf{F}, \mathbf{G}) \) from the FrontierNet outputs $(\mathbf{D}, \mathbf{Y})$, as defined by Eqs. \ref{eq:1} and \ref{eq:2}:
\begin{align}
    \mathbf{F}[x, y] &= 
    \begin{cases} 
      1 & \text{if } \mathbf{D}[x, y] < l \\ 
      0 & \text{otherwise} 
    \end{cases} \\
    G[i, j] &= \text{bin}^{-1}(\mathbf{Y}[i, j], K), \
    \label{eq:binning}
\end{align}
where \( l \) is the inclusion parameter for $\mathbf{F}$, and \( \operatorname{bin}^{-1}(\cdot, K) \) reverses the binning in \ref{eq:bin} to the lower bound of the bin. 

\begin{figure}
\centering
  \centering
  \includegraphics[scale=0.089]{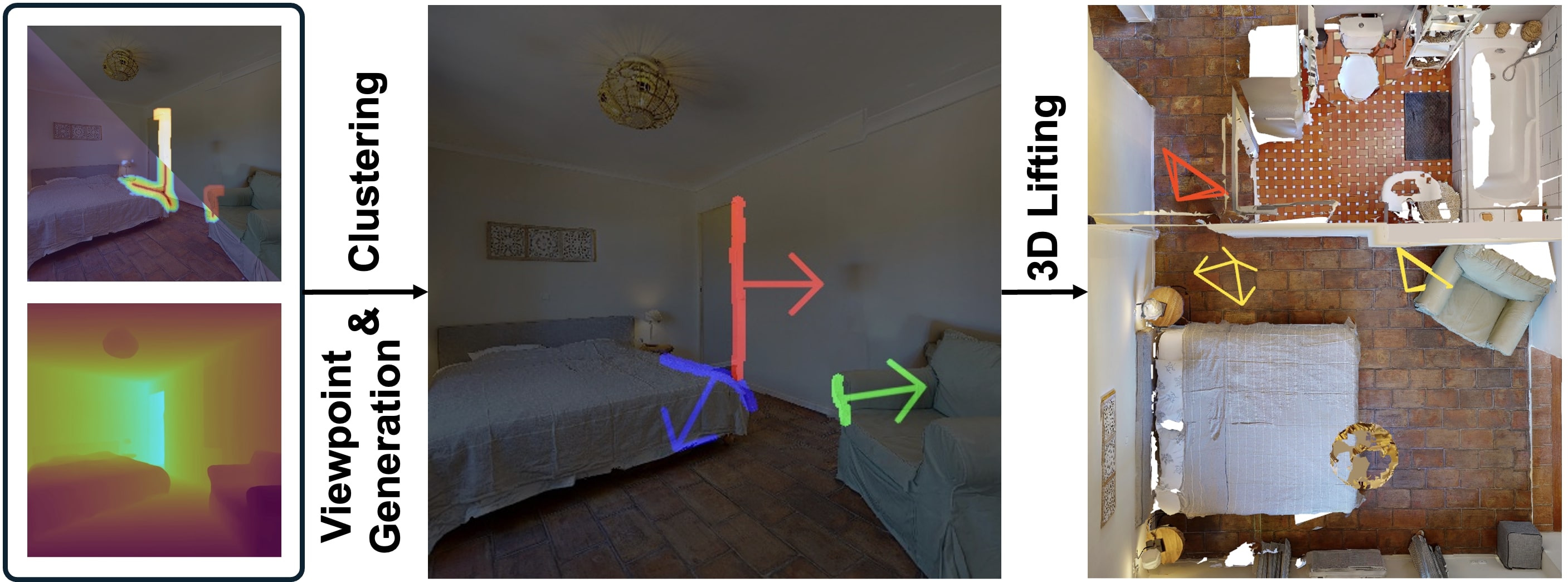}
    \vspace{-4.5mm}
\caption{\textbf{3D Frontier Generation. } Each frontier pixel is assigned a 2D viewing angle derived from the depth gradient. Combined with the info gain, 2D clustering is applied to obtain sparse 2D frontier clusters with associated viewing directions (middle). The foreground and background depths near the frontier pixels are then utilized to lift each clustered 2D frontier into 3D space (right). }
\label{fig:post}
\vspace{-1.5\baselineskip}
\end{figure} 

Fig. \ref{fig:post} then illustrates how \((\mathbf{F}, \mathbf{G})\) is converted into a set of sparse candidate viewpoints in the three‑dimensional scene through three successive steps: \emph{viewpoint generation}, \emph{clustering}, and \emph{3D lifting}.

\subsubsection{Viewpoint Generation}
\label{subsec:vp}

Viewpoint selection is often achieved through sampling-based approaches \cite{dai2020fast, papatheodorou2023finding, schmid2021unified, schmid2020efficient, jiang2024ag}. Our viewpoint generation method leverages monocular depth priors, eliminating the need for sampling operations in 3D. For each frontier pixel \((x, y)\), namely \( \mathbf{F}[x, y] = 1 \), we determine a 2D viewing direction from the depth gradient in its neighborhood. The gradient points along the steepest depth increase, typically from foreground to background. The gradient's inverse points toward the occluded space behind the foreground, providing the viewing direction \(\phi_{(x,y)}\) for \((x, y)\).

\subsubsection{Clustering}
3D frontier-based methods typically perform clustering on dense frontier voxels \cite{dai2020fast,tao2023seer,zhou2021fuel}. Similarly, we cluster 2D frontier pixels. We construct a feature vector \( \mathbf{Ft^{2D}} = [(x, y), \phi_{(x,y)}, g_{(x,y)}] \) for each frontier pixel. Here, \( \phi_{(x,y)} \) is again the viewing angle, and \( g_{(x,y)} = \mathbf{G}[x,y] \) is the info gain at \( (x, y) \). We cluster these feature vectors with HDBSCAN\,\cite{campello2013density} and obtain a sparse set of two–dimensional frontier clusters
\(\mathbf{Ft}^{\text{2D}}_{i},\; i=1,2,\dots,k\).
For each cluster we compute the representative feature \( [(\bar{x}_i, \bar{y}_i), \bar{\phi}_i, \bar{g}_i] \):\vspace{-0 pt}
\begin{itemize}[leftmargin=*]\setlength{\parskip}{0pt}
    \item The cluster coordinate \( (\bar{x}_i, \bar{y}_i) \) is the centroid pixel of its member pixels, ensuring it lies within the frontier pixels.
    \item The cluster’s viewing direction \( \bar{\phi}_i \) is the weighted average of the viewing directions of its member pixels, with weights assigned based on each pixel’s info gain.
    \item The cluster’s info gain \( \bar{g}_i \) is the average of all member pixels.
\end{itemize}

\subsubsection{3D Lifting}
\begin{figure}[t]
\centering
  \centering
  \includegraphics[scale=0.07]{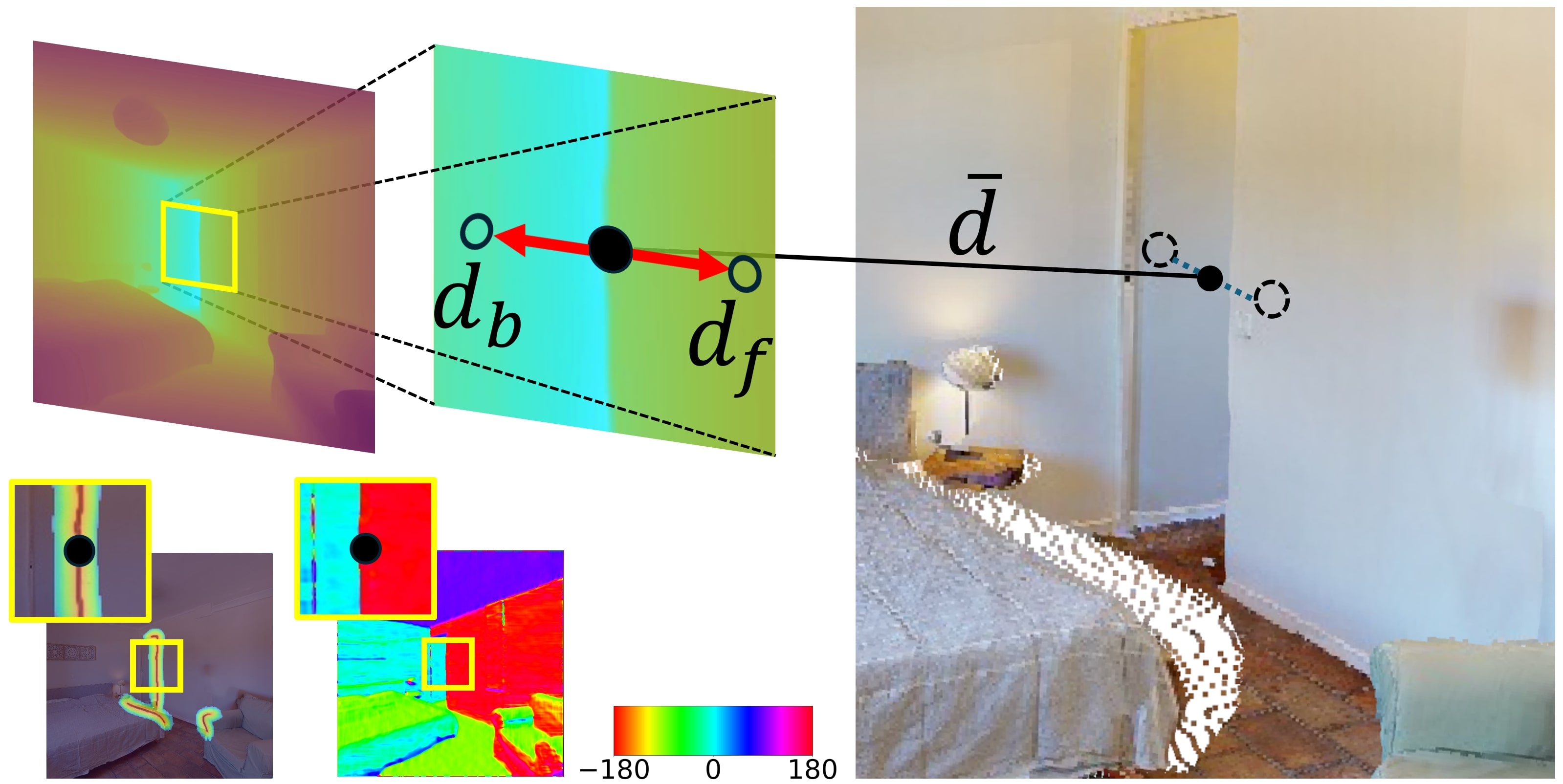}
  \vspace{-.5mm}
\caption{\textbf{Viewpoint Generation and 3D Lifting. } Our method computes a gradient map (bottom right) from the depth map. For each frontier pixel, foreground and background depths are sampled along the positive and negative gradient directions. The negative gradient also defines the 2D viewing angle, while the average of the two depths is used for lifting the pixel to 3D. }
\label{fig:lift}
\vspace{-1.4\baselineskip}
\end{figure}

To position the 2D frontiers in 3D, we assign each frontier pixel \( (x, y) \) a depth that lifts it to an intermediate location between the foreground and the background of the frontier. The lifting process begins with the same gradient map derived from the depth image as in viewpoint generation \ref{subsec:vp}. Two depth values, \( d_\text{b} \) and \( d_\text{f} \), are sampled along the positive and negative directions of the local depth gradient to approximate the depth of the background and foreground, respectively. The depth of the frontier is then calculated as the average, \( \bar{d} = (d_\text{b} + d_\text{f})/2 \). Fig. \ref{fig:lift} provides an example of this lifting operation for a single pixel. Although using the depth prediction in the process may not provide the exact metric depth everywhere, these errors in depth in this process are robustly compensated since: a) This approximation reliably captures the free space between the foreground and background, ensuring robustness against depth inaccuracies. b) To further enhance robustness, the depth values are assigned before clustering, and the final depth of each clustered frontier is taken as the average depth of its member pixels. Once the depth value for \( \mathbf{Ft}_{i}^\text{2D} \) is determined, its 3D viewpoint is obtained by lifting \( \bar{\phi}_i \) using the same depth value.

The entire anchoring process outputs a set of sparse 3D frontiers: $\mathbf{{Ft}}_{i}^{\text{3D}} = [\bar{\mathbf{p}}_i, \bar{\mathbf{q}}_i, \bar{g}_i]$ for $i = 1, 2, \dots, k$, 
where $\bar{\mathbf{p}}_i$ and $\bar{\mathbf{q}}_i$ represent the 3D position of frontier and the orientation of its viewing direction, 
and $\bar{g}_i$ denotes its info gain.

\subsection{Exploration Planning}

\subsubsection{Frontier Update}
Our system incorporates three primary update mechanisms for managing 3D frontiers, which operate concurrently as the robot explores.

\textbf{New Frontier Integration:} As new frontiers are proposed and lifted to 3D, they are added to the current 3D frontier list as new entries or merged with existing ones. Merging occurs because the same frontiers can be registered multiple times in 3D when viewed from different images capturing the same region. Following a similar metric used in the literature, each new frontier's 3D position and viewing direction are compared to those of all existing frontiers. 

If both the distance of the positions and the angle between the orientations of the new frontier and an existing frontier are below a threshold, the two are merged, with the properties of the merged frontier computed as the average of the two. Otherwise, the new frontier is registered independently. Since the 3D frontiers are sparse, this merging process remains computationally efficient, even as the list expands.

\textbf{Info Gain Adjustment:} Although our system extracts frontiers without relying on a 3D map, we can optionally maintain a 3D occupancy map to to refine frontier updates and to support safer path planning, both of which benefit from richer geometric context. Specifically, the initial info gain, \( \bar{g}_i \), of a frontier \( \mathbf{{Ft}}_{i}^\text{3D} \) reflects the unknown volume it can potentially observe without any information of the explored region. As the robot progresses the exploration, \( \bar{g}_i \) is expected to decrease. To capture this reduction, we project the known voxels \( \mathbf{v} \in \mathbf{V}_{\text{known}} \) from the current occupancy map into the image frame of \( (\bar{\mathbf{p}}_i, \bar{\mathbf{q}}_i) \), discard out‑of‑view or distant
projections, creating the set \( \mathbf{V}^{i}_{\text{known}} \). The updated info gain for \( \mathbf{{Ft}}_{i}^{\text{3D}} \) is then computed as: $\bar{g_i}' = \bar{g}_i - |\mathbf{V}^{i}_{\text{known}}|$.

\textbf{Invalid Frontier Removal:} A frontier \( \mathbf{{Ft}}_{i}^\text{3D} \) is considered invalid based on two criteria: a) if the system builds a 3D map, it checks if its updated info gain \( \bar{g_i}' \) falls below a minimum threshold \( g^{\text{min}} \), or b) if its viewpoint is similar to previously visited poses. This implies that the additional region indicated by a frontier has already been explored, or the frontier itself has been visited. To enforce the second criterion, we compare the Euclidean distance of positions and relative angle between its pose, \((\bar{\mathbf{p}}_i ,\bar{\mathbf{q}}_i)\), and the poses of the downsampled robot trajectory. This second criterion is especially important when info gain \( \bar{g}_i \) is inaccurately high in ambiguous scenarios, allowing such frontiers to be effectively cleared.

\subsubsection{Path Planning}

\begin{figure}
\centering
  \centering
  \includegraphics[scale=0.095]{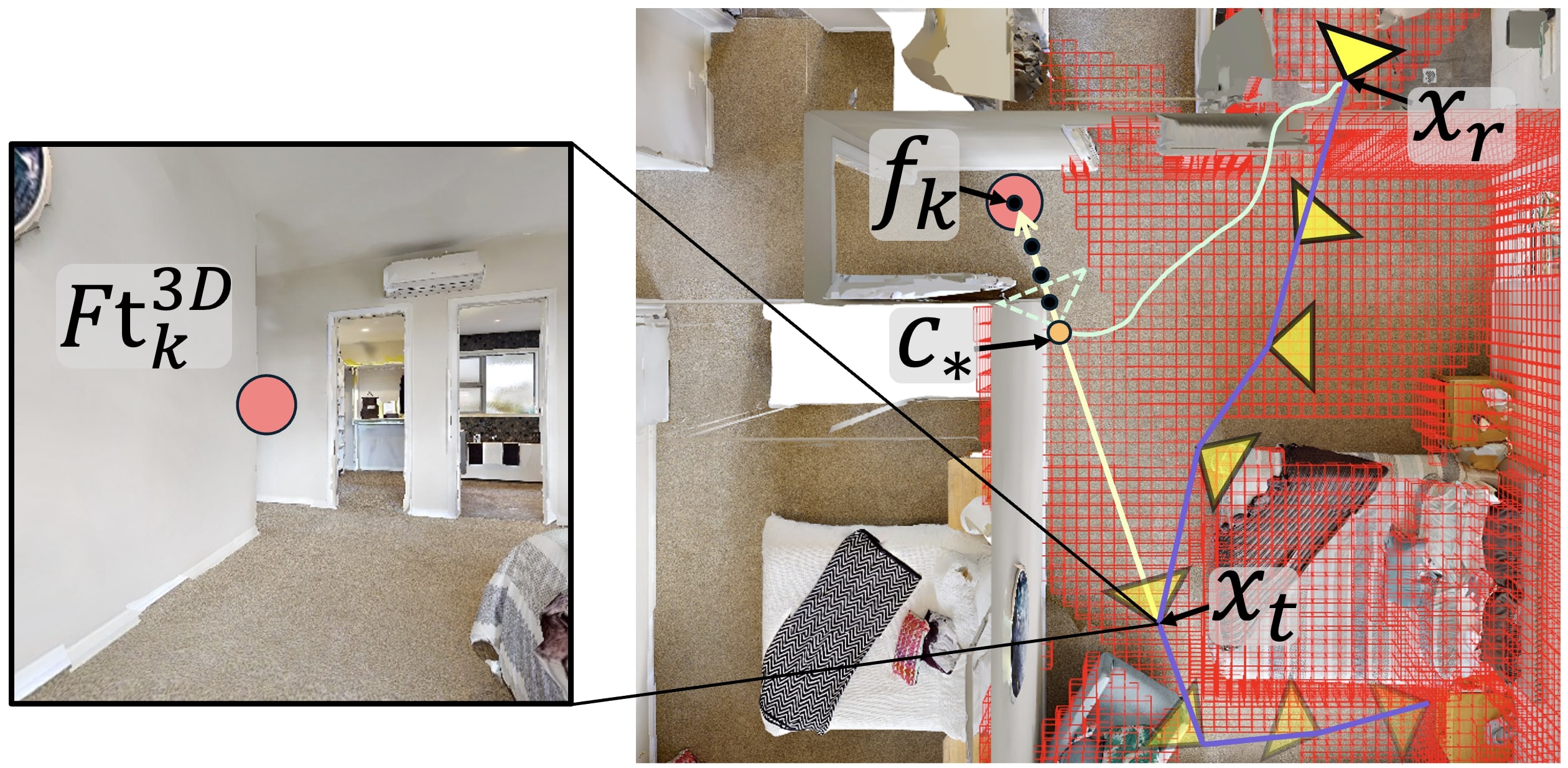}
  \vspace{-1.5mm}
\caption{\textbf{Path Planning. }
When the robot is at pose \(\mathbf{x}_r\), the next goal frontier \(\mathbf{f}_k\), proposed and registered by its previous pose \(\mathbf{x}_t\), lies outside the current 3D occupancy map (red voxelgrid). The planner samples points (black dots) backward along the edge \(( \mathbf{x}_t, \mathbf{f}_k)\) until it finds the nearest point \(\mathbf{c_{*}}\) within the map. The robot then plans a path to first navigate to \( \mathbf{c}_{*} \), and then incrementally map the surroundings while advancing toward \(\mathbf{f}_k\), using the edge as a directional prior.
}
\label{fig:plan}
\vspace{-1.5\baselineskip}
\end{figure}

Our path planning approach leverages frontier utility $u$ to guide the robot’s exploration. Similar to \cite{schmid2020efficient}, the utility of a candidate frontier \( \mathbf{{Ft}}_{i}^\text{3D} \) is defined as its info gain divided by the distance required to reach it:
\vspace{-0.5\baselineskip}
\begin{equation}
    u(\mathbf{x}_r, \mathbf{{Ft}}_{i}^\text{3D}) = \frac{\bar{g_i}'}{\|\mathbf{p}_{r}-\mathbf{p}_{i}\|},
\end{equation}

where \( \mathbf{x}_r = (\mathbf{p}_r, \mathbf{q}_r) \) is the current pose of the robot. The frontier with the highest utility is then selected as the next goal. This results in a balance between exploring nearby areas and pursuing more distant frontiers with potentially larger unknown regions, without additional tuning parameters.

During exploration, our planner maintains a rooted tree structure \( T = ({\mathcal{N}}, {\mathcal{E}}) \) that includes two types of nodes, \( {\mathcal{N}} = \{\mathbf{x}_0, \mathbf{x}_1, \mathbf{x}_2, \dots, \mathbf{x}_n, \mathbf{f}_1, \mathbf{f}_2, \dots, \mathbf{f}_m \} \), where \( \mathbf{x}_{(\cdot)} \) represents robot poses, and \( \mathbf{f}_{(\cdot)} \) denotes poses of valid frontiers. The nodes \( \mathbf{x}_{(\cdot)} \) form the main branch of the tree as a single chain: \( \mathbf{x}_0 \rightarrow \mathbf{x}_1 \rightarrow \mathbf{x}_2 \rightarrow \dots \rightarrow \mathbf{x}_n \). If the robot registers a frontier \( \mathbf{f}_j \) at a pose \( \mathbf{x}_i \), then \( \mathbf{f}_j \) is assigned to \( \mathbf{x}_i \) as its child, creating an edge \( (\mathbf{x}_i, \mathbf{f}_j) \in \mathcal{E} \). This frontier-camera linkage is a key feature enabled by FrontierNet, which proposes frontiers at the boundary between known and unknown regions, ensuring they are always within the camera's field of view and directly visible. This guarantees that at least one ray connects the camera's optical center to each clustered frontier. Consequently, the edge between the parent robot pose, and its frontier children represent both visibility and feasible traversability from the robot's pose to the frontier. Fig. \ref{fig:plan} illustrates a planning example of our system. When the next goal frontier \( \mathbf{f}_k \) lies beyond the current 3D map, our planner samples 3D points \( \mathbf{c}_{(\cdot)} \) along the edge to its parent robot pose \( \mathbf{x}_t = \operatorname{parent}(\mathbf{f}_k) \), verifying whether each sampled point \( \mathbf{c}_{(\cdot)} \) exists within the current occupancy map. Upon identifying the first valid point \( \mathbf{c}_{*} \) within the map, it is able to perform 3D path planning to reach \( \mathbf{c}_{*} \). To ultimately reach \( \mathbf{f}_k \), the robot uses the direct line between \( \mathbf{c}_{*} \) and \( \mathbf{f}_k \) as a prior and performs 3D path planning along this route while it maps more regions ahead.

Our planning approach is especially useful and reliable when the goal frontier lies far away or when inaccuracies arise due to scale differences in monocular depth estimation. Furthermore, it supports path planning in extreme scenarios, such as when computational or storage resources are limited or when depth sensing or prediction is highly unreliable, by enabling exploration solely with visibility information from the visual-only input.

\section{Experiment And Result}

\newcommand{\voxp}{Vol.}
\newcommand{\voxpt}{Vol@30}
\newcommand{\voxps}{Vol\@60}
\newcommand{\voxph}{Vol\@100}
\newcommand{\objp}{Obj.}
\newcommand{\succr}{Suc.}

\begin{table*}[t]
\setlength{\tabcolsep}{0.05cm}
\setlength{\tabcolsep}{0.25cm} 
\resizebox{\linewidth}{!}{
\begin{tabular}{l|l|cccccccccc|c}
\toprule
 &  & \multicolumn{1}{c}{824} & \multicolumn{1}{c}{827} & \multicolumn{1}{c}{876} & \multicolumn{1}{c}{880} & \multicolumn{1}{c}{804} & \multicolumn{1}{c}{807} & \multicolumn{1}{c}{812} & \multicolumn{1}{c}{834} & \multicolumn{1}{c}{854} & \multicolumn{1}{c|}{879} &  \\
\cmidrule{3-12}
 &  & {\scriptsize 10/79/21} & {\scriptsize 8/65/19} & {\scriptsize 14/148/8} & {\scriptsize 11/70/16} & {\scriptsize 10/111/11} & {\scriptsize 14/256/12} & {\scriptsize 8/67/16} & {\scriptsize 10/90/13} & {\scriptsize 6/72/5.0} & {\scriptsize 15/126/28} & \multicolumn{1}{c}{Mean} \\
\midrule
\multirow{5}{*}{\rotatebox{90}{Vox@25}} & $\mdwhtcircle$ Classic\cite{yamauchi1997frontier} &  $17.1\smallerpm{3.7}$  & $24.7\smallerpm{7.1}$ & $16.9\smallerpm{6.1}$ & $14.5\smallerpm{3.4}$ & $21.6\smallerpm{0.0}$ & $7.6\smallerpm{0.0}$ & $18.7\smallerpm{3.7}$ & $23.7\smallerpm{7.6}$ & $19.5\smallerpm{3.1}$ & $20.3\smallerpm{7.0}$ & 18.5\\
 &$\mdwhtcircle$ NBVP\cite{schmid2020efficient} & $23.4\smallerpm{3.6}$ & $21.6\smallerpm{4.1}$ & $23.3\smallerpm{4.4}$ & $18.9\smallerpm{1.8}$ & $24.5\smallerpm{3.5}$ & $17.5\smallerpm{3.3}$ & $27.2\smallerpm{3.0}$ & $23.4\smallerpm{6.5}$ & $26.9\smallerpm{3.2}$ & $22.5\smallerpm{5.1}$ & 22.9 \\
 & $\mdwhtcircle$ SEER\cite{tao2023seer} & $23.3\smallerpm{7.4}$ & $27.0\smallerpm{4.0}$ & $20.4\smallerpm{6.4}$ & $24.9\smallerpm{12.0}$ & $23.9\smallerpm{7.2}$ & $\times$ & $25.0\smallerpm{6.2}$ & $17.0\smallerpm{9.5}$ & $\bestres{34.7\smallerpm{1.6}}$ & $25.0\smallerpm{0.8}$ & 24.6 \\
 & $\odot$ SEER & $27.3\smallerpm{4.4}$ & $21.5\smallerpm{4.1}$ & $18.6\smallerpm{6.1}$ & $32.3\smallerpm{4.3}$ & $26.1\smallerpm{4.0}$ & $20.5\smallerpm{4.1}$ & $22.4\smallerpm{4.2}$ & $16.4\smallerpm{4.8}$ & $25.2\smallerpm{2.6}$ & $27.0\smallerpm{4.5}$ & 23.7 \\
 
 & $\mdwhtcircle$ Ours & $\bestres{32.2\smallerpm{3.7}}$ & \secondbest{$33.4\smallerpm{5.5}$} & $\bestres{33.5\smallerpm{4.1}}$ & \secondbest{$41.3\smallerpm{7.3}$} & \secondbest{$26.5\smallerpm{4.1}$} & \secondbest{$30.0\smallerpm{2.6}$} & $\bestres{38.3\smallerpm{7.5}}$ & $\bestres{30.6\smallerpm{3.0}}$ & \secondbest{$28.6\smallerpm{2.4}$} & $\bestres{32.2\smallerpm{5.0}}$ & $\bestres{32.7}$\\
 &$\mdblkcircle$ Ours & \secondbest{$31.3\smallerpm{4.7}$} & $\bestres{34.2\smallerpm{3.3}}$ & \secondbest{$31.6\smallerpm{4.2}$} & $\bestres{43.5\smallerpm{6.8}}$ & $\bestres{29.0\smallerpm{4.7}}$ & $\bestres{32.1\smallerpm{2.9}}$ & \secondbest{$37.0\smallerpm{9.5}$} & \secondbest{$29.7\smallerpm{2.8}$} & $27.9\smallerpm{4.4}$ & \secondbest{$30.7\smallerpm{5.8}$} & $\bestres{32.7}$\\
\midrule
\multirow{5}{*}{\rotatebox{90}{Vox@50}} &$\mdwhtcircle$ Classic & $29.1\smallerpm{4.8}$ & $37.6\smallerpm{8.0}$ & $31.9\smallerpm{7.2}$ & $26.1\smallerpm{7.2}$ & $39.4\smallerpm{0.0}$ & $24.2\smallerpm{0.0}$ & $27.7\smallerpm{6.7}$ & $37.5\smallerpm{5.6}$ & $43.1\smallerpm{4.4}$ & $39.2\smallerpm{5.6}$ &33.6 \\
 &$\mdwhtcircle$ NBVP & $46.2\smallerpm{5.7}$ & $46.1\smallerpm{5.9}$ & $44.5\smallerpm{5.1}$ & $31.0\smallerpm{1.3}$ & $46.6\smallerpm{4.6}$ & $35.3\smallerpm{2.7}$ & $49.4\smallerpm{6.6}$ & $44.1\smallerpm{3.0}$ & $52.3\smallerpm{2.6}$ & $45.5\smallerpm{4.8}$ &44.1\\
 &$\mdwhtcircle$ SEER & $42.1\smallerpm{5.7}$ & $42.7\smallerpm{6.3}$ & $36.5\smallerpm{11.4}$ & $49.5\smallerpm{5.5}$ & $35.6\smallerpm{9.3}$ & $\times$ & $47.3\smallerpm{4.1}$ & $24.4\smallerpm{16.8}$ & $46.1\smallerpm{4.0}$ & $43.6\smallerpm{4.1}$ &40.9 \\
 & $\odot$ SEER & $47.0\smallerpm{4.4}$ & $46.6\smallerpm{6.2}$ & $30.4\smallerpm{9.9}$ & $57.1\smallerpm{2.2}$ & $40.4\smallerpm{7.7}$ & $32.2\smallerpm{6.0}$ & $41.0\smallerpm{5.5}$ & $22.8\smallerpm{5.2}$ & $43.5\smallerpm{3.1}$ & $44.8\smallerpm{3.9}$ & 40.6 \\
 &$\mdwhtcircle$ Ours & $\bestres{58.0\smallerpm{4.8}}$ & $\bestres{61.9\smallerpm{3.9}}$ & $\bestres{58.2\smallerpm{4.2}}$ & $\bestres{61.9\smallerpm{7.5}}$ & \secondbest{$53.9\smallerpm{4.2}$} & $\bestres{50.7\smallerpm{4.5}}$ & \secondbest{$60.3\smallerpm{8.1}$} & \secondbest{$53.7\smallerpm{5.0}$} & $\bestres{72.1\smallerpm{9.8}}$ & \secondbest{$55.5\smallerpm{5.7}$} & $\bestres{58.6}$\\
 &$\mdblkcircle$ Ours & \secondbest{$56.6\smallerpm{7.2}$} & \secondbest{$60.1\smallerpm{6.2}$} & \secondbest{$51.0\smallerpm{8.6}$} & \secondbest{$60.9\smallerpm{4.9}$} & $\bestres{54.6\smallerpm{3.4}}$ & \secondbest{$45.4\smallerpm{4.4}$} & $\bestres{60.7\smallerpm{7.6}}$ & $\bestres{55.3\smallerpm{5.4}}$ & \secondbest{$53.5\smallerpm{6.6}$} & $\bestres{57.1\smallerpm{3.2}}$ & \secondbest{55.5}\\
\midrule
\multirow{5}{*}{\rotatebox{90}{Vox@100}} &$\mdwhtcircle$ Classic & $47.6\smallerpm{1.6}$ & $61.2\smallerpm{8.6}$ & $45.0\smallerpm{8.2}$ & $61.3\smallerpm{5.2}$ & $53.7\smallerpm{0.0}$ & $45.2\smallerpm{0.0}$ & $68.6\smallerpm{10.9}$ & $48.3\smallerpm{5.0}$ & $54.1\smallerpm{3.7}$ & $50.5\smallerpm{5.4}$ & 53.6\\
 &$\mdwhtcircle$ NBVP & $65.0\smallerpm{5.6}$ & $\bestres{78.5\smallerpm{4.9}}$ & $60.8\smallerpm{9.3}$ & $49.8\smallerpm{1.6}$ & $\bestres{69.7\smallerpm{4.8}}$ & $49.9\smallerpm{2.1}$ & $\bestres{83.4\smallerpm{3.5}}$ & \secondbest{$70.0\smallerpm{8.8}$} & $80.1\smallerpm{20.3}$ & $\bestres{62.6\smallerpm{5.6}}$ & 67.0\\
 &$\mdwhtcircle$ SEER & $55.6\smallerpm{5.1}$ & $50.7\smallerpm{5.0}$ & $51.0\smallerpm{8.6}$ & $54.0\smallerpm{3.8}$ & $56.6\smallerpm{4.1}$ & $\times$ & $54.8\smallerpm{7.7}$ & $44.2\smallerpm{3.0}$ & $48.9\smallerpm{6.8}$ & $50.3\smallerpm{2.5}$ & 51.8\\
& $\odot$ SEER & $60.6\smallerpm{6.7}$ & $60.1\smallerpm{5.6}$ & $50.5\smallerpm{8.8}$ & $60.3\smallerpm{6.1}$ & $62.3\smallerpm{3.2}$ & $51.7\smallerpm{5.6}$ & $60.8\smallerpm{8.3}$ & $45.1\smallerpm{4.9}$ & $51.0\smallerpm{3.4}$ & $48.1\smallerpm{3.0}$ & 55.1 \\
 &$\mdwhtcircle$ Ours & \secondbest{$71.2\smallerpm{6.0}$} & $72.6\smallerpm{8.9}$ & \secondbest{$72.0\smallerpm{8.5}$} & \secondbest{$68.4\smallerpm{10.8}$} & \secondbest{$62.2\smallerpm{8.9}$} & $\bestres{59.8\smallerpm{6.1}}$ & \secondbest{$82.2\smallerpm{10.1}$} & $\bestres{70.3\smallerpm{10.1}}$ & $\bestres{98.3\smallerpm{13.2}}$ & $58.8\smallerpm{6.5}$ &$\bestres{71.5}$\\
 &$\mdblkcircle$ Ours & $\bestres{73.0\smallerpm{8.5}}$ & \secondbest{$73.9\smallerpm{6.6}$} & $\bestres{72.7\smallerpm{9.0}}$ & $\bestres{70.9\smallerpm{9.3}}$ & $59.5\smallerpm{6.0}$ & \secondbest{$57.7\smallerpm{6.8}$} & $80.1\smallerpm{9.0}$ & $69.9\smallerpm{11.4}$ & \secondbest{$85.1\smallerpm{18.5}$} & \secondbest{$62.1\smallerpm{5.2}$} & \secondbest{70.6}\\
\midrule
\multirow{5}{*}{\rotatebox{90}{Suc.}} &$\mdwhtcircle$ Classic & 33.3 & \secondbest{86.7} & 38.0 & 40.0 & 6.3 & 5.6 & 37.5 & 31.3 & 90.0 & 20.0 & 38.9 \\
 &$\mdwhtcircle$ NBVP & $\bestres{100.0}$ & $\bestres{100.0}$ & $\bestres{90.0}$ & 50.0 & $\bestres{100.0}$ & 65.0 & $\bestres{100.0}$ & $\bestres{100.0}$ & 60.0 & $\bestres{100.0}$ & \secondbest{86.5}\\
 &$\mdwhtcircle$ SEER & 60.0 & 50.0 & 31.0 & 50.0 & 20.0 & 0.0 & 80.0 & 13.3 & 80.0 & 20.0 & 40.4\\
 &$\odot$ SEER & 88.9 & 55.6 & 61.1 & 66.7 & 55.6 & 33.3 & 55.6 & 33.3 & 80.0 & 77.8 & 60.8\\
 &$\mdwhtcircle$ Ours & $\bestres{100.0}$ & 81.3 & \secondbest{83.3} & $\bestres{100.0}$ & \secondbest{80.0} & $\bestres{80.0}$ & $\bestres{100.0}$ & 86.7 & $\bestres{100.0}$ & 75.0 &$\bestres{88.6}$ \\
 & $\mdblkcircle$ Ours & $\bestres{100.0}$ & 68.8 & 80.0 & \secondbest{90.0} & \secondbest{80.0} & \secondbest{75.0} & $\bestres{100.0}$ & \secondbest{90.9} & $\bestres{100.0}$ & \secondbest{80.0} & \secondbest{86.5} \\
\midrule
\end{tabular}
}
\vspace{-1.5mm}
\caption{\textbf{Quantitative Comparison. }Comparison of mapping efficiency (Vox@k\%) and success rate (Suc.) with baseline methods. Methods marked with an unfilled dot $\mdwhtcircle$ use ground-truth depth from the simulator, with a filled dot $\mdblkcircle$ use metric monocular depth estimation \cite{hu2024metric3d}. $\odot$ SEER is our re‑implementation of the original frontier‑proposal technique, paired with our planner and evaluated under identical test conditions with perfect depth. The 3-digit numbers in the first row are scene IDs. The three parameters below each scene ID are the retrieved relevant scene parameters from HM3D metadata (num\_rooms, navigable\_area, and navigation\_complexity).}
\label{table:quanti}
\vspace{-6mm}
\end{table*}

\subsection{Experiments Setup}

Our exploration system is evaluated on validation scenes from HM3D that were never used to train either FrontierNet or the monocular depth estimator. The chosen scenes span a wide range of sizes and geometric complexity. We simulate camera viewpoints and render images with Open3D~\cite{zhou2018open3d}. Without loss of generality, image has a resolution of \( 480 \times 480 \). The field of view (FOV) angle and the maximum depth range of the sensor are set to \({77.32^\circ \times 77.32^\circ}\) and \({3.5}\)m.  Depth input is provided in two variants: (1) perfect depth rendered from the scan and (2) depth predicted by Metric3D v2\cite{hu2024metric3d}. We employ a Python wrapper of Octomap \cite{hornung13auro} to build the occupancy map. Our low-level 3D path planner is implemented using the Open Motion Planning Library (OMPL) \cite{sucan2012the-open-motion-planning-library}.

For quantitative evaluation, camera motion is simulated by interpolating the planned path into dense, discrete poses and steering the camera through these poses. We benchmark our method against a classic frontier method~\cite{yamauchi1997frontier}, a more recent approach, SEER~\cite{tao2023seer}, and a sampling-based approach,
NBVP~\cite{schmid2020efficient}. No open-source code is available for \cite{yamauchi1997frontier}, so we implement it ourselves. We use the official ROS implementations for \cite{schmid2020efficient} and \cite{tao2023seer} to get the exploration paths. Additionally, we include a SEER variant that replaces our FrontierNet with SEER’s frontier proposal while inheriting the rest of our pipeline.

Autonomous exploration lacks a standardized test protocol, and most methods are generally tested on a limited number of scenarios. Specifically, the two recent baselines selected for the benchmarking were merely evaluated in two scenes. To measure both efficiency and generalization, we expand the test scenarios to 10 diverse scenes, varying in layout, size, appearance, and number of floors. In every scene, we place the camera at several initial poses and let it explore until either no frontiers remain or a scene‑specific step limit is reached. Each start pose is repeated five times. A new step is registered when the robot undergoes a significant change in position (\( >0.1\, \text{m} \)) or orientation (\( >10^\circ)\), ensuring travel distance in both translation and rotation are considered. To accommodate these large and varied testbed, we also introduce an evaluation metric that is comparable between environments of varying size and complexity: \textbf{Vox@}$k$(\%), the fraction of total scene volume explored when the number of steps reaches \( k \%\) of the total steps. To evaluate exploration efficiency across stages, we report \( k = 25, 50, 100 \). To determine the statistical steps for \( k=25 \) and \(50\) for each scene, we calculate the average step count across all methods at which \( 25\% \) and \( 50\% \) volume coverage is reached. This average step threshold is applied to each method individually, measuring the volume coverage achieved at this common step count. This approach ensures interpretability by reflecting the expected performance at a consistent stage across methods, without favoring any specific approach. For \( k = 100 \), the step count corresponds to the maximum threshold during exploration. A trial is successful if it achieves Vox@$100$ \( > 40\% \). From this, we compute the average success rate, \textbf{Suc.}(\%).
\vspace{-3mm}
\subsection{Result}
We conduct experiments to investigate several questions:

\textbf{How does our approach’s exploration efficiency compare with baselines?}

\begin{figure}
\centering
  \includegraphics[scale=0.101]{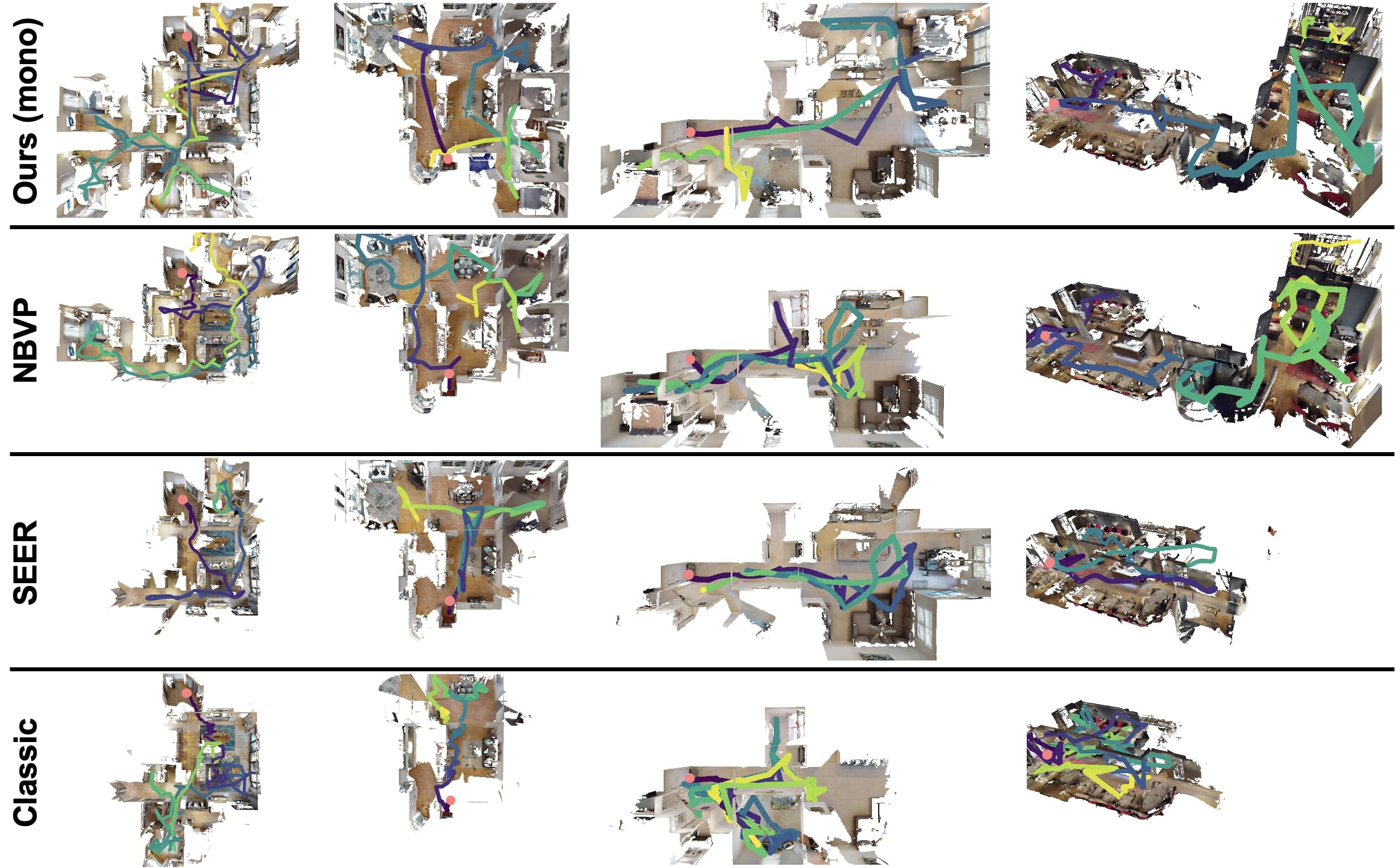}
  \vspace{-5mm}
\caption{\textbf{Qualitative Comparison. }Exploration examples of our method (using predicted depth) compared to three baseline methods (using perfect depth) across four scenes (left to right: 876, 824. 880, 854). Starting location is marked as red point. Notably, our approach successfully handles multi-floor environments (scene 854), a challenge for traditional frontier-based methods. All 3D meshes in this visualization are generated by TSDF integration using ground-truth depth images just for fair and clearer comparison.}
\label{fig:qual_result}
\vspace{-1.5mm}
\end{figure}

Table \ref{table:quanti} summarizes the quantitative results of our experiments. Across all 10 scenes, our method with simulator depth input consistently achieves the highest overall efficiency at 25\%, 50\%, and 100\% of total steps, as well as the highest success rate. Our method using a monocular depth prior ranks second in these metrics, performing better than baseline methods with simulator depth. Notably, at Vox@25, our method outperforms baseline approaches in nine scenes, and at Vox@50, it surpasses all baselines in all 10 scenes, exceeding the second-best method by around 15\% overall. This demonstrates the ability of our system to effectively prioritize regions with higher info gain during early exploration. It is important to note that all baseline methods rely on simulator depth to ensure accurate 3D maps for extracting goal poses to explore. Switching to monocular depth estimation would significantly degrade their performance, as inaccurate maps that caused by depth scale errors or artifacts lead to failures in generating feasible goal poses. In contrast, our method maintains robust performance even with monocular depth inputs. Fig. \ref{fig:qual_result} provides qualitative examples of this experiment. We use the same path planner, frontier assignment, and update logic for our method, our implementation of the Classic method, and SEER. FrontierNet’s superior results therefore arise solely from its own strengths: it detects frontiers more reliably and estimates information gain more accurately. These improvements show that leveraging the visual cues leads to more effective exploration.

\textbf{How do the RGB and depth images individually contribute to the performance of FrontierNet?}

To explore this, we train multiple FrontierNet models with different input configurations: RGB-only, depth-only, and RGB\&depth. We then compare the model performance on a validation set. As shown in Table \ref{table:abl1}, results indicate that both color and depth information are essential for accurate distance field detection and info gain estimation. Specifically, detection relies predominantly on geometric cues from depth, whereas info gain estimation benefits from both appearance cues from RGB and geometric cues from depth as we have hypothesized. 

\begin{table}[t]
\centering
\setlength{\tabcolsep}{0.75em} 
\begin{tabular}{lrrr}
\toprule
& \multicolumn{1}{c}{RGB-only} & \multicolumn{1}{c}{Depth-only} & \multicolumn{1}{c}{RGB\&Depth} \\
\midrule
Distance Field Err. \scriptsize{(pixels)} $\downarrow$ & 0.315 & 0.167 & \textbf{0.152} \\
Info Gain Cls. Dice Score $\uparrow$ & 0.406 & 0.403 & \textbf{0.440} \\
\bottomrule
\end{tabular}
\vspace{-1mm}
\caption{\textbf{Performance of FrontierNet Models with Different Inputs.}}
\label{table:abl1}
\vspace{-6mm}
\end{table}

\textbf{How much does the distance field and info gain map contribute to the final exploration efficiency?} 

\begin{figure}
\vspace{-7mm}
\centering
  \centering
  \includegraphics[width=0.49\linewidth]{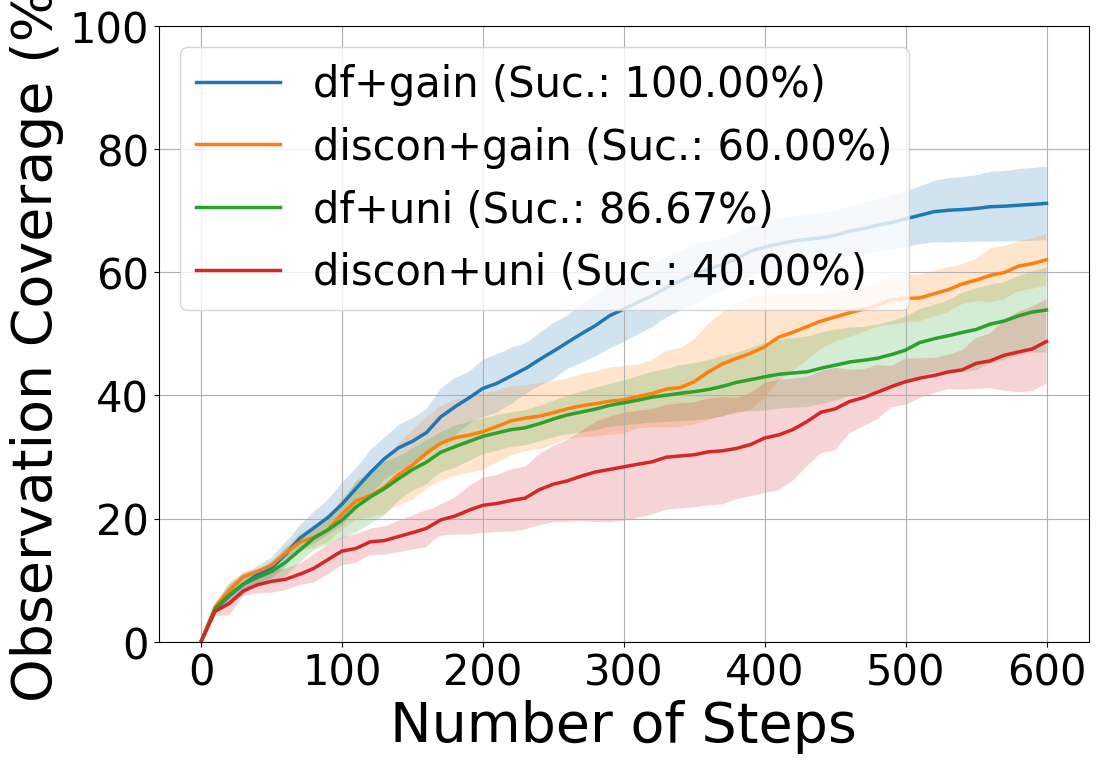}
  \includegraphics[width=0.49\linewidth]{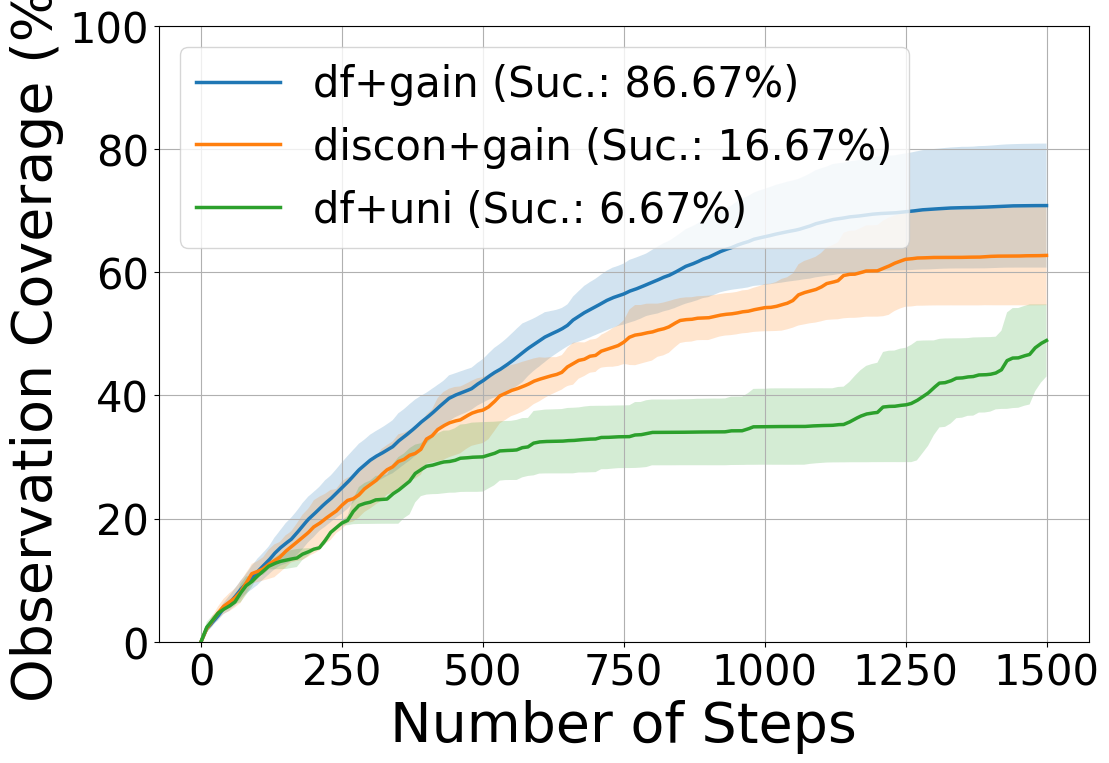}
  \vspace{-6mm}
\caption{Performance comparison of different configurations on scene 824 (left) and scene 876 (right). The discon+uni configuration completely fails on scene 876 (0.0\% Suc).}
\label{fig:abl2}
\vspace{-1.5\baselineskip}
\end{figure}

We select two scenes and perform exploration with different planner configurations: (1) \textit{df+gain}: using the predicted distance field and info gain; (2) \textit{df+uni}: using the predicted distance field with uniform info gain, assigning the same gain to all pixels; (3) \textit{discon+gain}: using the depth discontinuity mask along with info gain. This mask, identical to $\mathbf{F}_\text{d}$ in Fig. \ref{fig:gt_gen}, is extracted from the input depth; (4) \textit{discon+uni}: using the discontinuity mask with uniform info gain. We track the percentage of mapped volume achieved by each configuration.

As shown in Fig. \ref{fig:abl2}, efficiency and success rate drop when the planner lacks either accurate frontier pixels or info gain. Without info gain, it treats all frontiers equally, leading to suboptimal paths prioritizing nearby frontiers. Without distance field detection, the discontinuity mask generates a noisy, redundant map boundary, adding significant overhead to the process. The results confirm that efficient exploration depends on both the distance field and the info gain, and that predicting them with a learned model provides a clear advantage.

\textbf{How does our system perform in a fully map-free setup?}

\begin{figure}
\centering
  \centering
  \includegraphics[scale=0.065]{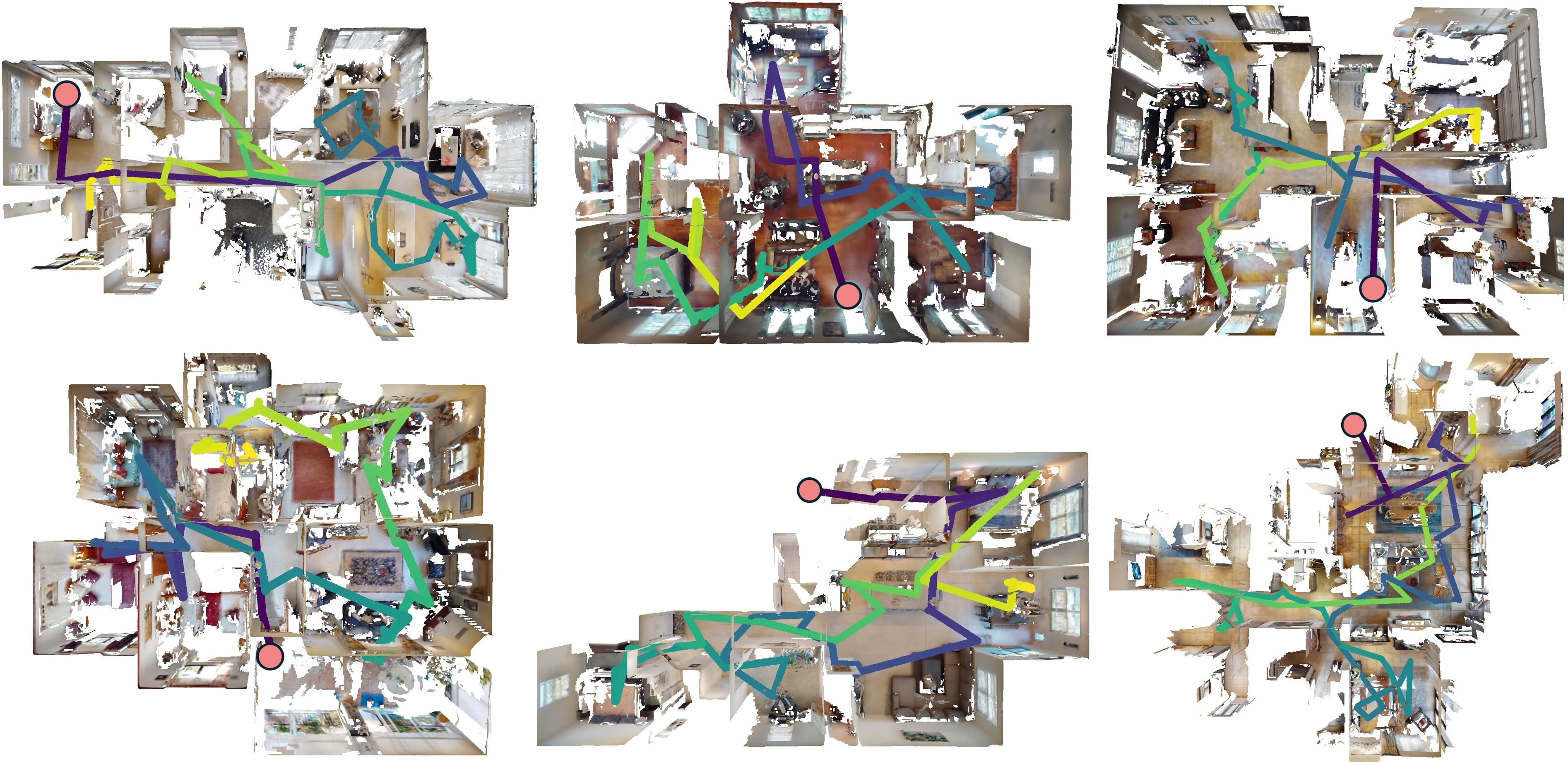}
  \vspace{-1.0mm}
\caption{\textbf{Map-Free Exploration Example.} Examples across six different scenes (Top row: scenes 804, 827, 879; bottom row: scenes 883, 880, 876.) No dense 3D map is maintained during exploration; the reconstructions shown serve only as visualizations.}
\label{fig:topo}
\end{figure}

FrontierNet proposes frontiers from visual input alone, and both the frontier update and planning modules can run without a dense 3D map. In this configuration the system relies solely on the frontier tree and the robot’s past trajectory. The loss of the map mainly affects path planning and the info gain adjustment: the robot keeps only the second validation rule, which checks whether a frontier lies close to a pose it has already visited and thus serves as a sparse memory. To examine the concept we evaluated this map-free setting in six scenes with predicted depth. Table \ref{table:mapfree} and Fig. \ref{fig:topo} show that the robot still achieves promising results, with little revisiting behaviour and performance comparable to the baselines. These findings suggest that our approach can function entirely without geometric maps and can be extended to tasks such as object search or goal directed navigation.

\begin{table}[t]
\setlength{\tabcolsep}{0.1cm}
\centering
\scriptsize
\begin{tabular}{l|cccccc|c}
\toprule
& 804 & 827 & 876 & 879 & 880 & 883 & Mean \\
\midrule
Vox@25  & 20.2 & 24.1 & 23.9 & 25.4 & 25.1 & 21.0 & 23.3 \\
Vox@50  & 36.6 & 40.6 & 38.5 & 37.2 & 39.8 & 37.7 & 38.4 \\
Vox@100 & 52.7 & 59.3 & 57.2 & 50.7 & 55.6 & 60.2 & 56.0 \\
\midrule
\end{tabular}
\vspace{-1.5mm}
\caption{\textbf{Map-free Exploration Result.} Performance across six HM3D scenes with predicted depth. For each scene, a subset of the initializations (3 out of 5) used to get results in Table~\ref{table:quanti} is sampled. Reported Vox@k\% metrics follow the same.}
\label{table:mapfree}
\vspace{-2.5mm}
\end{table}

\subsection{Real-world Validation}

We implement our exploration system as a ROS package and deploy it on a Boston Dynamics Spot robot. A calibrated camera in the front provides \( 640 \times 480 \) RGB images at 3 Hz. Our software runs on a laptop with an i9-12900HX, 32 GB RAM, and a 16 GB 3080Ti GPU. FrontierNet achieves  $\sim 5$ Hz inference, enabling real-time image processing. 

\begin{figure}
\centering
  \centering
  \includegraphics[scale=0.066]{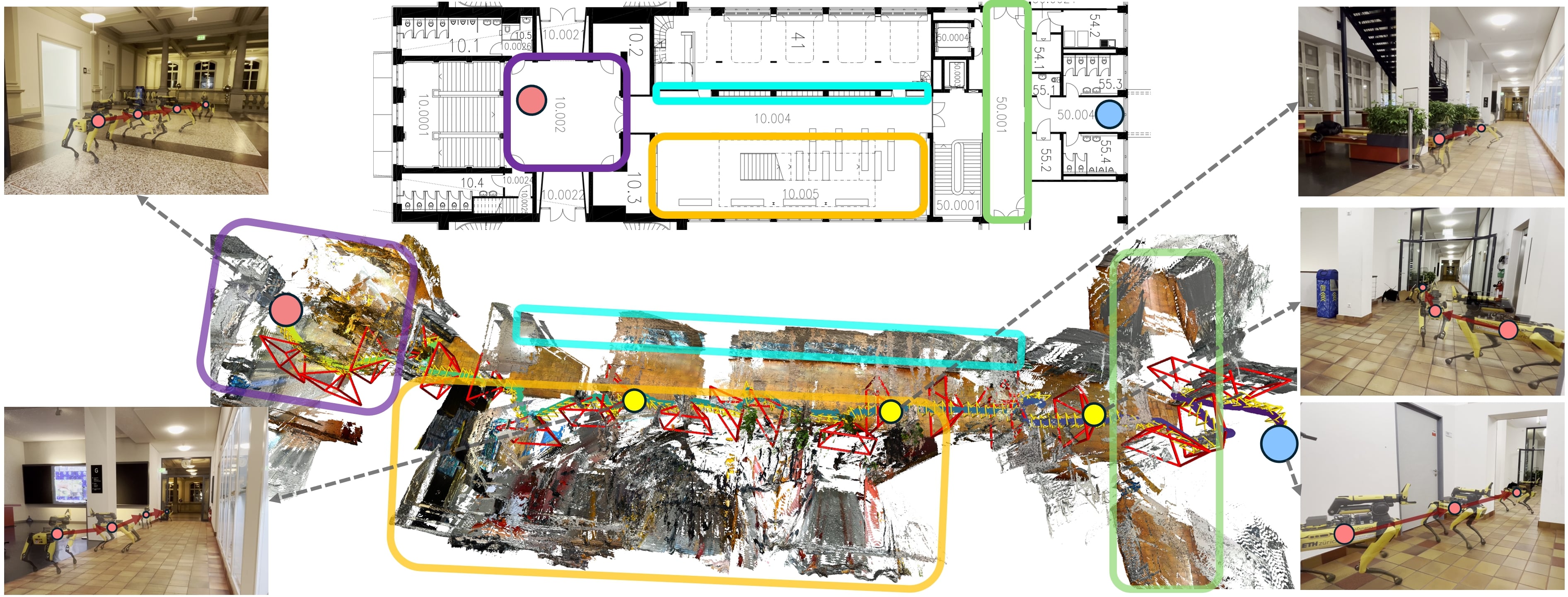}
\vspace{-.5mm}
\caption{\textbf{Real-world Validation Result. }Exploration process of a quadrupedal robot in a real-world environment. Top: Floor plan. Bottom: Reconstructed map and exploration path from TSDF integration using monocular depth prediction. Colored boxes indicate key correspondences between the map and floor plan.}

\label{fig:spot}
\vspace{-1.5\baselineskip}
\end{figure}

Fig. \ref{fig:spot} shows the exploration process in a large indoor environment. Despite being trained solely on renderings, FrontierNet demonstrates strong robustness to the sim-to-real gap. The robot successfully maps cluttered corridors, always prioritizes large, unexplored space, and ultimately reaches the far side of the entrance without human intervention.

\section{Conclusion}

In this work, we investigate how to leverage both appearance and geometric information from visual input to enable efficient autonomous exploration. We propose FrontierNet, a hybrid model for 2D frontier proposal and information gain prediction, and design an exploration system to integrate seamlessly with it. Our system demonstrates significant advantages in exploration efficiency without relying on a 3D map to generate exploration goals. We validate its effectiveness through extensive simulation and real-world experiments.
\vspace{-0.4\baselineskip}


\bibliographystyle{IEEEtran}
\bibliography{references}

\end{document}